%% file: Main.tex
\documentclass[a4paper,fleqn]{cas-sc}

\usepackage{float}

\usepackage{pythonhighlight}
\usepackage{tabularx}
\usepackage[english]{babel}
\usepackage[utf8]{inputenc}
\usepackage{amsmath,amssymb}
\usepackage{setspace}
\usepackage{graphicx}
\usepackage{wrapfig}
\usepackage{subfigure}
\usepackage{multicol}
\usepackage{multirow}
\usepackage{todonotes}

\usepackage{lscape}
\usepackage{pdflscape}
\usepackage{algpseudocode}
\usepackage{algorithm}      

\usepackage{xcolor}

\usepackage[authoryear]{natbib}
\def\tsc#1{\csdef{#1}{\textsc{\lowercase{#1}}\xspace}}
\tsc{WGM}
\tsc{QE}

\begin{document}
\let\WriteBookmarks\relax
\def\floatpagepagefraction{1}
\def\textpagefraction{.001}

\shorttitle{Traffic Weather Graph CGNs for EV Charging Forecasting}

\shortauthors{J. Tupayachi et al.}

\title [mode = title]{Spatio-Temporal Graph Convolutional Networks for EV Charging Demand Forecasting Using Real-World Multi-Modal Data Integration}

\author[1]{Jose Tupayachi}
\cormark[1]                                
\fnmark[1]                                 
\ead{jtupayac@vols.utk.edu}               
\credit{Conceptualization, Software, Data Curation, Investigation, Writing – original draft, Visualization, Writing – review \& editing}

\author[1]{Mustafa C. Camur}
\ead{mcamur@utk.edu}
\credit{Supervision,  Writing – review \& editing, Project Administration}

\author[1]{Kevin Heaslip}
\ead{kheaslip@utk.edu}
\credit{Project Administration}


\author[1]{Xueping Li}
\ead{xli27@utk.edu}
\credit{Supervision, Writing – review \& editing, Funding Acquisition}

\affiliation[1]{x
    organization={University of Tennessee},
    department={Department of Industrial and Systems Engineering},
    addressline={Knoxville, TN}, 
    country={USA}
}



\cortext[1]{Corresponding author}

\fntext[1]{Email: jtupayac@vols.utk.edu; tupayachisja@ornl.gov}


\begin{abstract}
Transportation remains a major contributor to greenhouse gas emissions, highlighting the urgency of transitioning toward sustainable alternatives such as Electric Vehicles (EVs). Yet, uneven spatial distribution and irregular utilization of charging infrastructure create challenges for both power grid stability and investment planning. This study introduces Traffic-Weather Graph Convolutional
Network (TW-GCN), a spatio-temporal forecasting framework that combines Graph Convolutional Networks with temporal architectures to predict EV charging demand in Tennessee, United States. We utilize  real-world  traffic flows, weather conditions, and proprietary data provided by one of the largest U.S.-based EV infrastructure companies  to capture both spatial dependencies and temporal dynamics. Extensive experiments across varying forecasting horizons, clustering strategies, and sequence lengths reveal that mid-horizon (3-hour) forecasts achieve the best balance between responsiveness and stability, with One-dimensional convolutional neural networks consistently outperforming other temporal models. Regional analysis shows disparities in predictive accuracy across East, Middle, and West Tennessee, reflecting how station density, Points of Interest and local demand variability shape model capabilities. The proposed TW-GCN framework advances the integration of data-driven intelligence into EV infrastructure planning while supporting sustainable mobility transitions.
\end{abstract}


\maketitle











\input{content}










\bibliographystyle{cas-model2-names}

\bibliography{cas-refs}






\end{document}

%% file: content.tex
\section{Introduction}

Transportation is an indispensable part of human life, and with the exponential growth of the urban population, its importance continues to increase. In 2023, transportation services in the United States (U.S.) contributed 8.9\% to the gross domestic product. Adjusted for inflation, the demand for transportation increased by 5.5\% in 2023. Despite its economic benefits, transportation also imposes several negative externalities, including traffic congestion and environmental pollution.

Greenhouse gas (GHG) emissions, a key driver of climate change and global warming, are primarily caused by the consumption of carbon fuels (e.g., gasoline) during energy use. In 2022, the transportation sector was reported as one of the two main contributors to total CO$_2$ emissions \citep{raimi2022global}. The transportation sector accounted for approximately 28\% of total U.S. GHG emissions in 2022 \citep{EPA2025}. Subsequently, in 2023, the transportation sector accounted for about 37\% of total U.S. energy consumption~\citep{EIA_US_Energy_Facts_2024}.

Electric vehicles (EVs), together with shared mobility, play a vital role in the decarbonization and electrification of transportation networks~\citep{cui2022battery}. \textcolor{black}{As of January 2025, the U.S. had approximately 196{,}000 publicly available EV charging stations~\citep{ClimateCentral2025_EVChargeUp}.} To support this transition, the Bipartisan Infrastructure Law allocated~\citep{FHWA2025evcharging} \$2.5~billion through the Discretionary Grant Program for Charging and Fueling Infrastructure and the National EV Infrastructure Formula Program, including a recent \$635~million in grants to further expand public EV charging nationwide.

Nevertheless, the growing adoption of EVs results in an imbalance between supply and demand, especially at commercial charging stations. In certain regions, high utilization of EV charging stations exerts substantial stress on power grids \citep{roy2023impact}. In contrast, underutilization in  other areas can cause inefficient resource allocation and financial losses for investors and operators \citep{etxandi2023estimation}. Therefore, understanding the behavior of EV charging demand is crucial. This involves not only analyzing the spatial distribution of EV stations but also considering factors such as temporal features, weather features, urban geographical features, and traffic-based data. 



To address this problem, scholars have explored a variety of approaches, such as probabilistic models \citep{tang2015probabilistic}, time series analyses \citep{amini2016arima}, machine learning algorithms \citep{liu2018hybrid}, and deep learning techniques \citep{wang2023short}. These demand prediction methods can be categorized into two groups: (a) long-term demand prediction (e.g., yearly) and (b) short-term demand prediction (e.g., hourly). Short-term predictions are particularly beneficial for dynamic price adjustments and managing load demands, while long-term predictions aid in infrastructure development and urban planning.

To the best of our knowledge, no prior study has analyzed real-world U.S.-based (i.e., Tennessee) EV charging demand data while simultaneously accounting for spatial, weather, and traffic flows derived from TMCs metrics. In this study, we focus on Tennessee, a key state in the Southeastern U.S., where detailed datasets of charging station activity, traffic flow, and weather conditions are available. Using these real-world operational datasets, we aim to forecast energy consumption (in kWh) at EV charging stations. To achieve this, we propose a novel Traffic-Weather Graph
Convolutional Network  (TW-GCN) model that integrates historical traffic and weather time series within a stacked graph convolutional network (GCN) architecture. This model captures both temporal and spatial dependencies to predict the charging demand. Our multidimensional approach identifies clusters of stations with similar usage patterns to deliver region-specific, accurate EV charging demand forecasts that can inform both infrastructure planning and mobility transitions.



The paper is organized as follows. Section~\ref{Literature Review} reviews existing work on EV station demand prediction and model architectures. Section~\ref{Data_discussion} describes the datasets, preprocessing, and quality checks. Section~\ref{Machine Learning Methodology} outlines the model architecture, training, and evaluation metrics. Section~\ref{Computational Experiments } presents results, model comparisons, and statistical analyses. Section~\ref{Discussion} focuses on a Tennessee case study across short, mid, and long-term horizons. Finally, Section~\ref{Conclusion} presents the conclusions and outlines directions for future research.

\section{Literature Review} \label{Literature Review}

In this section, we review the existing studies from two perspectives. First, we discuss the evolution of Spatial-temporal Graph Neural Networks and our reasoning for choosing them as our solution technique (Section~\ref{Emergence}). Next, we examine state-of-the-art methods for EV charging demand prediction (Section~\ref{EV-Charging}), where the increasing availability of historical data has led many researchers to employ deep learning approaches. Finally, we outline our contributions in Section~\ref{Contribution}.

\subsection{Emergence of Spatial-temporal Graph Neural Networks \& Graph Convolutional Networks}
\label{Emergence}
In recent years, the increasing availability of spatial-temporal data has led to the development of advanced deep learning models capable of capturing both spatial dependencies and temporal dynamics. Convolutional Neural Networks (CNNs) have demonstrated considerable efficiency, particularly when handling grid-like data structures such as images and spreadsheets \citep{li2021survey}. However, many real-world data structures are irregular, including but not limited to protein–protein interaction networks, transportation networks, and infrastructure networks. To address this issue, Recurrent Neural Networks (RNNs) have been employed by many researchers \citep{salehinejad2017recent}. Yet, they face certain challenges, such as vanishing and exploding gradients, difficulties in learning long-term dependencies, and restrictions related to sequential data processing, such as time series.


To overcome these limitations, Graph Neural Networks (GNNs) have been proposed to extend convolutions to graph data structures; see \citet{wu2020comprehensive} for an extensive review. The term GNN serves as an umbrella concept encompassing a broad family of architectures designed to process graph-structured data. Within this family, several specialized models have been developed, such as GCNs, Graph Attention Networks (GATs), and Graph Recurrent Networks (GRNs), each introducing distinct mechanisms for information propagation and aggregation across nodes. Initial studies in this field predominantly focused on spectral representations based on the eigenvalues and eigenvectors of the graph Laplacian matrix (i.e., $L = A - D$, where $A$ is the adjacency matrix and $D$ is a diagonal matrix whose entries $D_{ii}$ represent the degree of node $i$) to perform graph convolutions \citep{bruna2013spectral}.

Due to the computational burden associated with these spectral methods, research later shifted toward spatially localized filters \citep{berg2017graph}.
However, as stated by \citet{velickovic2017graph}, the learned filters from these methods do not generalize well to different graph structures in terms of node and edge connections. This limitation led to the development of non-spectral approaches such as GraphSAGE, which creates node embeddings through sampling and aggregating features from local neighborhoods \citep{hamilton2017inductive}. A significant advancement was made with the introduction of GATs by \citet{velickovic2017graph}, which combine self-attention mechanisms with convolutional methods to address the node classification problem.
Incorporating spatial-temporal dimensions into GNNs has further enhanced their applicability, particularly in dynamic networks where nodes and edges change over time. Spatial-temporal GNNs (ST-GNNs) integrate spatial dependencies between nodes with temporal evolution \citep{yu2017spatio}. 

ST-GNNs have been successfully employed in traffic flow forecasting \citep{ali2022exploiting}, traffic accident forecasting \citep{yu2021deep}, demand prediction \citep{wang2023predicting}, and passenger demand prediction \citep{tang2021multi}. These models are particularly well-suited for charging demand forecasting since the problem is inherently spatial-temporal.
EV charging demand is influenced not only by time-dependent usage cycles (e.g., commute times, weekdays versus weekends, seasonal variations), but also by spatial correlations between stations that arise from road connectivity, geographic clustering of demand, and heterogeneous adoption patterns. Conventional forecasting models such as regression, time series analysis, or deep learning architectures often treat space and time independently, which limits their ability to capture cross-location spillover effects and evolving network interactions. In contrast, ST-GNNs are explicitly designed to learn from both spatial topologies and temporal dynamics, making them more effective at modeling when and where demand will arise. Building on these strengths, we propose a novel approach to forecast the demand for EV charging in the Southeastern U.S. (i.e., the Tennessee region).

\subsection{EV Charging Demand Forecasting} \label{EV-Charging}

The interest in EV charging station demand forecasting has been growing exponentially. Early studies primarily focused on EV charging station load, which involves forecasting the electrical load measured in kilowatts or megawatts \citep{qian2010load}. In one of the earliest contributions, \citet{xie2011research} applied neural networks (NNs), including radial basis function variants, alongside a predictive model to estimate daily EV charging station loads using data from the 2010 Beijing Olympic Games site. Following this, \citet{xydas2013electric} analyzed a fleet of EVs to estimate the weekly charging demand in kW using decision tables, representative trees, NNs, and support vector machines (SVM). While these methods were effective in capturing temporal patterns of charging load, they often do not explicitly model spatial dependencies or interactions between different stations and external factors such as traffic or weather. To address this limitation, we adopt graph-based approaches that naturally incorporate both temporal and spatial relationships, motivating our choice of GNNs for the forecasting task.

In contrast, \citet{majidpour2014novel}  utilized a time series method along with Random Forest (RF), SVM, and k-Nearest Neighbor (kNN) to estimate energy consumption at the charging outlet level. However, the authors did not use any spatial features or driving habit-based data. The following year, \citet{olivella2015probabilistic} proposed an agent-based simulation model to understand the cumulative EV charging demand and its grid impact. One of the very first studies analyzed social (e.g., age) and economic (e.g., income) variables, which was further extended \citep{arias2016electric} in which real-world traffic and weather data were used to forecast EV charging demand; nevertheless, clustering and decision tree approach does not capture complex interactions between traffic, weather, and charging demand, particularly in dynamic scenarios or under high EV penetration scenarios.

Spatial-temporal charging demand prediction was conventionally tackled by simulation  and probabilistic models \citep{zhou2012spatial, zhang2015integrated}, then extended by using  dynamic probabilistic models \citep{xia2019ev}, where they considered driving behavior, battery usage, and traffic information. 
Moving forward, a short-term load forecasting model \citep{mcbee2020forecasting} using SVM was developed based on the European Commission’s Off Transport dataset, where the time of arrival, departure time, daily travel distance, and battery state of charge were considered to forecast the charging load. The authors also compared the proposed method with the Monte Carlo forecasting technique but did not account for critical factors like geographical location, weather, or traffic data. To address these limitations, \citet{kim2021forecasting} introduced an ensemble machine learning model including Autoregressive Integrated Moving Average (ARIMA), NNs, and  Long Short-Term Memory (LSTM) for forecasting EV charging demand. They aimed to enhance accuracy by considering factors such as charging patterns, weather conditions, and the day of the week. Further, the authors analyze three geographical scales: station-, city-, and country-based.

Other approaches have contemplated fuzzy sets, which are based on fixed-value rules, and cannot be optimized through learning from big data. Among these, \citet{zamee2023self} performed a Maximal Information Coefficient correlation to identify dominant features in asynchronously sampled datasets and applied a General Regression NN-based forecasting model. They selected dominant lagged exogenous variables through correlation multi-collinearity analysis. Although they claim to handle big data and feature selection by correlation, the sources of the big data, the pipeline, and the reasons why certain features were considered significant have not been discussed. To develop a predictable and economically interpretable model, \citet{kuang2024physics} proposed a learning approach for accurate EV charging demand prediction and rational pricing, called PIAST, which integrates convolutional feature engineering, a spatio-temporal dual attention mechanism, and physics-informed neural network training \citep{raissi2019physics}. They incorporated the concept of price elasticity of demand, which measures how responsive demand is to price changes. However, the model did not account for external factors such as weather and traffic density.

\citet{sun2021exploring} investigates the role of EVs in supporting the smart grid through vehicle-to-grid interactions. Using a one-month GPS trajectory dataset from 967 rental EVs in Beijing (January 2019), the authors applied fixed cutoff values to the adjacency matrices to analyze charging and discharging patterns. However, this static analytical approach failed to capture the evolving dynamics of demand patterns. Although the study highlights the importance of analyzing demand from rental EVs, which significantly contribute to the grid, it overlooks the combined impact of private and rental EVs on the overall load. Relying solely on rental GPS trajectories limited the analysis, as it failed to capture station-level demand variations influenced by geographic location, weather, or nearby amenities. Examining station demand evolution could reveal how such external factors shape charging behavior across regions.

Building on this limitation, \citet{yi2022electric} proposed a Seq2Seq model for forecasting monthly commercial EV charging demand. The authors complemented this with clustering techniques to capture regional spatiotemporal demand patterns. Yet clustering methods often treat data points in isolation, relying mainly on feature similarity. This limitation hinders their ability to capture richer contextual dependencies between locations; in contrast, GNNs can incorporate multiple layers of context while explicitly modeling interlocation relationships.

Recent studies have applied deep learning to EV charging demand forecasting, using approaches such as LSTMs for short-term station-level prediction~\citep{wang2023short} and GCNs combined with gated recurrent units to capture spatio-temporal features~\citep{wang2023predicting}. GCNs, in particular, have demonstrated strong performance in modeling spatial relationships between charging stations and users~\citep{huttel2021forecasting, wang2023gcn} for capturing the dynamic and interconnected nature of EV demand. To further improve forecasting accuracy, it is important to incorporate external factors such as Points of Interest (POI) data, weather conditions, and traffic density, which can reveal evolving non-linear patterns in station-level demand. Table~\ref{tab:lit_summary_compact} summarizes these studies, highlighting their methodological innovations and key limitations that motivate our proposed approach.

\begin{table}[htpb!]
\caption{Recent Studies in Spatiotemporal Forecasting for EV Charging Demand}
\label{tab:lit_summary_compact}
\tiny
\begin{tabularx}{\textwidth}{@{}p{1cm} p{3.5cm} p{4.8cm} p{3cm} X@{}}
\toprule
\textbf{Authors} & \textbf{Approach} & \textbf{Contribution} & \textbf{Limitations} & \textbf{Coverage} \\
\midrule

\citet{wang2025adaptive} & Adaptive Spatial-Temporal Graph Recurrent Network (ASTGRN) & Eliminates the need for predefined geographic distances or prior spatial knowledge & Does not incorporate traffic inputs and evaluated only on GRU architecture & Beijing, China \\

\citet{li2025spatiotemporal} & Non-negative Matrix Factorization (NMF) to extract latent spatiotemporal EV charging patterns from high-resolution driver data & Large-scale driver data revealed three main charging patterns: late night, daytime, and early evening & Excludes external factors such as weather, user behavior, and traffic & Four major cities in China \\

\cite{cui2025weather}  & Random Forest combined with SHAP values  & Identification of weather in origin–destination pairs to affect long-distance highway travel for EVs & Model may not generalize to colder regions, high-traffic periods, or areas with different infrastructure & Guangdong, China \\

\cite{huang2025energy} & Energy Consumption Transformer (ECT) & Addresses EV data scarcity using a shot-based learning approach with transferable prior knowledge. & Limited generalization to regions with distinct driving patterns or environmental conditions. & Shanghai, China \\

\cite{chen2025disaggregation} & Bi-level Spatio-Temporal Graph Convolutional Network (ST-GCN) framework. & Event and design dissagregation for non-intrusive community-level EV charging profile disaggregation & Tested on limited GRU, Transformer, and LSTM architectures, with no further benchmark analysis & Selected states in the United States \\

\cite{zheng2025spatio}  & Spatio-Temporal Charging Load Prediction Method based on Cellular Traffic Simulation (CTM) & Real-time traffic parameters alongside charging load predictions, allowing charging station planning and traffic-aware charging guidance  & Only analyses a single urban region & Hunan, China\\

\cite{shahrokhi2025two} & Two-Stage Clustering and Classification & Combines spatial (where) and temporal (when) charging behaviors to group EV users for accurate grid impact forecasting & Primarily focuses on temporal features, with limited consideration of spatial mobility and accessibility factors & Saskatchewan, Canada \\

\cite{wu2025spatio}  & Spatio-temporal prediction framework for EV charging load based on real road networks & Incorporates varied EV types and trip behavior to model energy consumption & Limited to a single city and simulation-based validation & Yunnan, China \\

\cite{tian2025multi}  & Multi-Scale Spatial-Temporal Graph Attention Network (MSSTGAN) & Uses graph attention-based gated fusion to dynamically integrate spatio-temporal embeddings, capturing the importance of different stations (spatial) and historical time points (temporal) & Evaluated on geographically dispersed networks without traffic or weather data & Palo Alto, Boulder (U.S.), Perth (U.K.) \\

\bottomrule
\end{tabularx}
\end{table}

\subsection{Our Contribution}
\label{Contribution}

Recognizing an existing research gap, this paper proposes a TW-GCN, which combines a GCN model for capturing spatial features of station-level geographic characteristics with a temporal model that processes historical time series data, including weather and traffic information. This stacked architecture integrates spatial and temporal predictors to forecast EV charging demand. Our contributions are threefold:

\begin{itemize}
\item We incorporate real-world traffic and weather features into a temporal GCN framework and analyze the impact on forecasting performance, benchmarking against baseline methods.
\item We leverage proprietary EV transaction data from a leading charging network operator in Tennessee to study charging behavior in a region characterized by slower yet steadily growing EV adoption.
\item We perform computational experiments that combine time series models with graph convolutional networks to improve the accuracy of EV charging demand predictions.
        
\end{itemize}

\section{Datasets} \label{Data_discussion}

To effectively model and predict charging station energy consumption, we compile and integrate diverse datasets capturing temporal, spatial, and contextual factors. We forecast energy consumption (kWh) at charging stations by leveraging these various influencing variables illustrated in Table~\ref{tab:data_features_with_definitions}. Temporal attributes~\citep{zhang2025high}, such as \texttt{start\_date\_time}, reveal usage patterns~\citep{senol2023electric, dominguez2020seasonality}, while geographic coordinates (\texttt{station\_latitude}, \texttt{station\_longitude}) capture spatial relationships~\citep{li2018planning}. Additionally, traffic-related metrics, weather conditions, and POI data are incorporated to provide a richer modeling context. These temporal, spatial, and contextual attributes reflect real-world charging behavior, motivating the use of a commercial dataset to analyze station usage and infrastructure patterns.

\begin{table}[htbp]
\centering
\tiny
\caption{Features and Definitions per category.}\label{tab:data_features_with_definitions}
\begin{tabular}{ll}
\toprule
\textbf{Feature} & \textbf{Definition} \\
\midrule

\multicolumn{2}{l}{\textbf{POI (Points of Interest)}} \\
\midrule
has\_supermarket & Indicates presence of a supermarket near the location \\
has\_retail\_shopping & Indicates presence of retail shopping areas nearby \\
has\_higher\_educ & Indicates presence of higher education institutions \\
has\_school & Indicates presence of schools \\
has\_park & Indicates presence of parks or recreational green space \\
has\_restaurant & Indicates presence of restaurants \\
has\_police & Indicates presence of a police station \\
has\_library & Indicates presence of a library \\
has\_hospital & Indicates presence of a hospital \\
\midrule

\multicolumn{2}{l}{\textbf{Traffic (Performance Metrics)}} \\
\midrule
speed & Current observed vehicle speed at the station \\
historical\_average\_speed & Average speed over a historical period \\
reference\_speed & Baseline speed for comparison purposes \\
speed\_deviation & Difference between observed and reference/historical speed \\
delay\_per\_mile & Delay per mile traveled (traffic delay indicator) \\
travel\_time\_seconds & Estimated travel time in seconds \\
charging\_time\_seconds & Estimated EV charging time in seconds \\
energy\_kwh & Energy consumption in kilowatt-hours \\
tti & Travel time index (traffic flow efficiency indicator) \\
cvalue & Quantitative indicator of congestion severity \\
confidence\_score & Confidence level in the prediction or measurement \\
is\_congested & Boolean indicating if the station is currently congested \\
\midrule

\multicolumn{2}{l}{\textbf{Weather}} \\
\midrule
pressure & Atmospheric pressure at the location \\
temperature & Air temperature in degrees Celsius \\
humidity & Relative humidity percentage \\
precip & Precipitation amount \\
wind\_speed & Wind speed in the area \\
\midrule

\multicolumn{2}{l}{\textbf{Temporal / Location}} \\
\midrule
start\_date\_time & Timestamp indicating start of the observation \\
station\_id & Unique identifier for the traffic station \\
station\_latitude & Latitude coordinate of the station \\
station\_longitude & Longitude coordinate of the station \\
\bottomrule
\end{tabular}
\end{table}

\subsection{ChargePoint Dataset \& Stations}

EV adoption has accelerated in recent years, fueling the expansion of charging infrastructure to meet growing demand. Reliable and comprehensive datasets on station usage are crucial for understanding user behavior, optimizing network deployment, and enhancing operational efficiency. The ChargePoint network, the largest independently owned EV charging provider in North America and Europe, operating in more than 14 countries, offers valuable insights into charging patterns and infrastructure utilization~\citep{investors2024globalreach}. As of January 2025, it accounts for 61\% of publicly available AC charging ports in North America and provides access to over one million charging locations when including roaming partners~\citep{sec2025filing}. In Tennessee, 86 ChargePoint stations represent about 7.5\% of the state’s 1,142 public charging locations~\citep{doe2025afdc}.

The dataset used in this study comprises detailed session-level logs from these stations over 13 months. Each log captures key variables such as charging duration, energy dispensed, and station-specific metadata. To support our modeling framework, the dataset is further enhanced with the weather, traffic, and points of interest (POI) features, enabling a more nuanced analysis of spatial and temporal demand patterns. 



\subsection{Points of Interest}

Understanding the factors that shape charging demand requires considering the built environment around stations. In particular, nearby POIs can significantly influence whether certain stations are used more frequently than others~\citep{pagany2019electric}. Using a minimum-distance criterion~\citep{straka2020predicting}, we incorporate POIs into various variables, including spatial patterns derived from POI proximity. This integration helps to capture spatial heterogeneity in charging demand. We selected locations within a 500-meter radius of each station to capture local demand patterns while maintaining spatial specificity, as shown in Table~\ref{tab:data_features_with_definitions}. A 500-meter buffer, commonly used in urban studies, captures the typical walkable distance for EV users seeking nearby amenities, enabling analysis of spatial factors influencing charging demand \citep{ren2025predicting}. While POI information reflects potential interactions with demand estimates, factors such as off-street parking availability may respond to local conditions like weather~\citep{lyu2024short}. In our study, we collected POI information from OpenStreetMap~\citep{OpenStreetMap}.

\subsection{Weather and Traffic Information}

Environmental factors such as weather and traffic conditions significantly influence EV charging behavior and station performance. In our work, we selected weather features including temperature, pressure, and humidity, as these directly affect the performance of the lithium-ion batteries used in EVs. For example, batteries experience reduced efficiency and accelerated degradation in cold environments~\citep{wang2025electric}, which can increase charging duration and, in turn, station usage. Conversely, hot climates, particularly in arid regions, create additional challenges for battery efficiency and station operations~\citep{el2023impact}.  

In addition, we considered traffic indicators such as average speed and congestion within the EV corridor. These factors capture the effect of bottleneck-constrained roads, where structural and dynamic interactions can significantly alter charging behavior~\citep{wang2021electric}.

Weather-related variables such as temperature, humidity, and precipitation were obtained from Weather Underground~\citep{WeatherUnderground}. Transportation-related variables were derived from Traffic Message Channels (TMCs), which encode real-time traffic events including congestion, accidents, and road closures using standardized location and event codes~\citep{iso14819, wright2008tmc}. These features enable dynamic rerouting and optimized traffic flow within EV corridors by incorporating up-to-date traffic patterns. The traffic data was sourced from the Regional Integrated Transportation Information System (RITIS)~\citep{ritis}.

\section{Proposed Methodology}\label{Machine Learning Methodology}

Studies~\citep{shahriar2021prediction,  huttel2021forecasting, chen2023hybrid} have examined methods to address the complex task of forecasting EV charging demand. These approaches commonly integrate contextual factors such as weather, traffic, and event data to predict session duration and energy consumption using traditional machine learning models. While incorporating temporal dependencies improves predictive robustness, most methods do not account for spatial relationships, even when temporal features are combined with traffic, weather, and POI information.

To capture the complex relationships between entities such as charging stations connected by traffic patterns or spatial proximity, \textit{graph-based learning} provides a natural framework. In this paradigm, each node $v \in \mathcal{V}$ in a graph $\mathcal{G} = (\mathcal{V}, \mathcal{E})$ is associated with a feature vector $\mathbf{x}_v \in \mathbb{R}^F$, where $F$ denotes the number of features. Here, $\mathcal{V}$ represents the set of nodes (e.g., charging stations), and $\mathcal{E}$ represents the set of edges that capture relationships or interactions (e.g., spatial proximity or traffic influence). These vectors capture key attributes of the entities represented by the nodes, enabling the model to incorporate both structural and contextual information. 

In our study, node features serve as the primary input to the Network and capture intrinsic attributes of charging stations, including geographic location, capacity, historical energy usage, and operational metrics like session duration and charger availability. These vectors enable the model to learn contextualized representations through iterative message passing and neighbor aggregation.
Node features vary by domain, for instance, voltage and power measurements in power grids~\citep{suri2025gnn}, or socio-demographic, land use, and traffic data in EV infrastructure modeling~\citep{batic2025evgnn}. Message passing layers iteratively update node features by incorporating information from adjacent nodes, allowing each station to reflect its local network context.

The GNN learns contextualized representations by iteratively updating node features through message passing from their local neighborhood. Formally, the updated feature \(\mathbf{x}_v^{(\ell+1)}\) of node \(v\) at layer \(\ell+1\) (Equation \ref{eq:passingmessage}) is computed as:

\begin{equation}\label{eq:passingmessage}
\mathbf{x}_v^{(\ell+1)} = f_{\theta}^{(\ell+1)} \left( \mathbf{x}_v^{(\ell)}, \{ \mathbf{x}_w^{(\ell)} : w \in \mathcal{N}(v) \} \right)
\end{equation}

Where \(\mathcal{N}(v)\) denotes the set of neighboring nodes of \(v\), individual vertex in the set of nodes \(\mathcal{V}\) of the graph \(\mathcal{G} = (\mathcal{V}, \mathcal{E})\) and \(f_{\theta}^{(\ell+1)}\) is a learnable function that aggregates the node features with those of its neighbors. We adopt the GCN formulation introduced by~\citet{kipf2016semi}, in which the layer-wise propagation rule (Equation \ref{eq:gcn}) is defined as:

\begin{equation}\label{eq:gcn}
\mathbf{X}^{(\ell+1)} = \sigma\left( \hat{\mathbf{D}}^{-\frac{1}{2}} \hat{\mathbf{A}} \hat{\mathbf{D}}^{-\frac{1}{2}} \mathbf{X}^{(\ell)} \mathbf{W}^{(\ell)} \right)
\end{equation}

Here, \(\hat{\mathbf{A}} = \mathbf{A} + \mathbf{I}\) is the adjacency matrix with added self-connections, \(\hat{\mathbf{D}}\) is the diagonal degree matrix of \(\hat{\mathbf{A}}\), \(\mathbf{W}^{(\ell)}\) is the learnable weight matrix at layer \(\ell\), and \(\sigma(\cdot)\) is a nonlinear activation function, such as ReLU. We adopt the standard notation in network analysis, where the open neighborhood of a node \( v \), denoted \( \mathcal{N}(v) \), refers to the set of nodes adjacent to \( v \), while the closed neighborhood, denoted \( \mathcal{N}[v] \), includes \( v \) itself in addition to its neighbors: $\mathcal{N}[v] = \mathcal{N}(v) \cup \{v\}$.

We model the system as a graph $\mathcal{G} = (\mathcal{V}, \mathcal{E})$, where: Nodes $v \in \mathcal{V}$ represent stations, while edges $(w,v) \in \mathcal{E}$ denote a connection from station $w$ to station $v$. The closed neighborhood of a node $v$, denoted $\mathcal{N}[v]$, is defined as the set containing the node itself along with all nodes $w$ that have an edge pointing to $v$; that is, $\mathcal{N}[v] = \{v\} \cup \{ w \mid (w,v) \in \mathcal{E} \}$.

\subsection{Node Update in Graph Neural Networks}

Let $\mathbf{x}_v^{(\ell)}$ be the feature vector of node $v$ at layer $\ell$. The update rule for a GNN at layer $\ell+1$ is: \(\mathbf{W}^{(\ell+1)}\) (Equation \ref{eq:nodeupdate}) is a trainable weight matrix, 
\(c_{w,v}\) is a normalization coefficient associated with edge \((w,v)\), 
and \(w\) is a source node sending information to the target node \(v\).

\begin{equation}
\label{eq:nodeupdate}
\mathbf{x}_v^{(\ell+1)} = \mathbf{W}^{(\ell+1)} \sum_{w \in \mathcal{N}[v]} \frac{1}{c_{w,v}} \mathbf{x}_w^{(\ell)},
\end{equation}

\subsection{Temporal Stacking}

Given $T$ time steps, the hidden representations for each adjacency type are stacked as tensors (Equation \ref{eq:stacking}):

\begin{equation}\label{eq:stacking}
\begin{aligned}
\mathbf{H}^{\text{geo}} &= \big[ \mathbf{H}_1^{\text{geo}}, \mathbf{H}_2^{\text{geo}}, \dots, \mathbf{H}_T^{\text{geo}} \big] \in \mathbb{R}^{T \times N \times H}, \\
\mathbf{H}^{\text{dem}} &= \big[ \mathbf{H}_1^{\text{dem}}, \mathbf{H}_2^{\text{dem}}, \dots, \mathbf{H}_T^{\text{dem}} \big] \in \mathbb{R}^{T \times N \times H}.
\end{aligned}
\end{equation}

where $N$ is the number of nodes (stations), $H$ is the hidden feature dimension, and $T$ is the number of time steps. To introduce the temporal components, the GCN outputs are passed to a time series model (e.g., LSTM). Represented by: $\mathbf{Z}^{\text{geo}} \in \mathbb{R}^{T \times N \times H}, \quad 
\mathbf{Z}^{\text{dem}} \in \mathbb{R}^{T \times N \times H}$

In this framework, variables are classified as either geographical or demand-based according to the following rule: time-varying quantities are incorporated into $\mathbf{Z}^{\text{dem}}$, while location-specific or static quantities are incorporated into $\mathbf{Z}^{\text{geo}}$. Specifically:

\begin{itemize}
    \item \textbf{Demand-related features} ($\mathbf{Z}^{\text{dem}}$): variables that evolve 
    over time and reflect activity on the network, such as vehicle counts. Weather descriptors (e.g., average annual temperature or 
    long-term climate indicators
    \item \textbf{Geographical features} ($\mathbf{Z}^{\text{geo}}$): variables that are tied 
    to a location and remain relatively static, such as 
    POIs. 
\end{itemize}

This separation ensures that $\mathbf{Z}^{\text{dem}}$ captures dynamic demand signals, 
while $\mathbf{Z}^{\text{geo}}$ encodes spatial context and structural characteristics. A weighted combination of the geographical and demand vectors is then taken:

\begin{equation}
\mathbf{Z} = \alpha_1 \mathbf{Z}^{\text{geo}} + (1 - \alpha_1) \mathbf{Z}^{\text{dem}},
\label{eq:alpha}
\end{equation}

where $\alpha_1 \in [0,1]$ balances the contribution of the geographic features $\mathbf{Z}^{\text{geo}}$ and the demand-related features $\mathbf{Z}^{\text{dem}}$. In this context, the variables can be associated with either component as follows: demand-related features (such as vehicle counts or congestion levels) are captured by $\mathbf{Z}^{\text{dem}}$, while static spatial characteristics (such as road network properties, weather as a location-tied factor, and nearby points of interest) are incorporated into $\mathbf{Z}^{\text{geo}}$.

To enhance forecasting, K-means clustering is applied to the features at the last time step \(T\), which groups nodes into a set of regions \(\mathcal{R} = \{1, 2, \ldots, K\}\), where \(K\) is the number of clusters. This clustering-based regionalization not only mitigates noise at the individual station level but also captures shared demand dynamics, leading to more stable and scalable forecasting compared to modeling each station independently. The K-means algorithm (Equation \ref{eq:region1}) assigns each node \(i\).
\begin{equation}
\label{eq:region1}
    \text{Region}(i) = \arg\min_{j \in \mathcal{R}} \| \mathbf{Z}_T(i) - \mathbf{C}_j \|,
\end{equation}

where the index \(j\) denotes the cluster or region identified by K-means, and \(\mathbf{C}_j\) is the centroid of the \(j\)-th cluster. For each node \(i\), this expression assigns it to the region \(j\) whose centroid \(\mathbf{C}_j\) is closest to the node’s feature vector \(\mathbf{Z}_T(i)\) at the last time step \(T\), effectively grouping nodes into regions based on feature similarity. An MLP is then applied to model the nonlinear relationship between the shared feature representation \(\mathbf{Z}\) and the energy consumption \(\hat{\mathbf{Y}}_i\) for each region, enabling cluster-specific predictive behavior. MLPs serve as fundamental components of state-of-the-art transformer architectures, where they operate alongside attention mechanisms to capture complex dependencies over learned representations. The final output applies a ReLU activation to ensure non-negative kWh values, aligning with physical constraints (Equation~\ref{eq:region2}).

\begin{equation}
\label{eq:region2}
\hat{\mathbf{Y}}_i = \mathbf{W}_{\text{Region}(i)} \mathbf{Z}
\end{equation}

where \( \mathbf{W}_{\text{Region}(i)} \)  is the weight matrix for region \( i \). The final output is represented via a ReLU activation, which ensures non-negative kWh values. To support and enrich region-specific predictions, the following section presents the creation and handling of graph structures based on Haversine distance and Dynamic Time Warping, capturing both spatial and temporal dependencies among regions.

\subsection{Graph Generation} \label{Graph_Generation}

\begin{figure}[!htpb]
    \centering
    \includegraphics[width=1\linewidth]{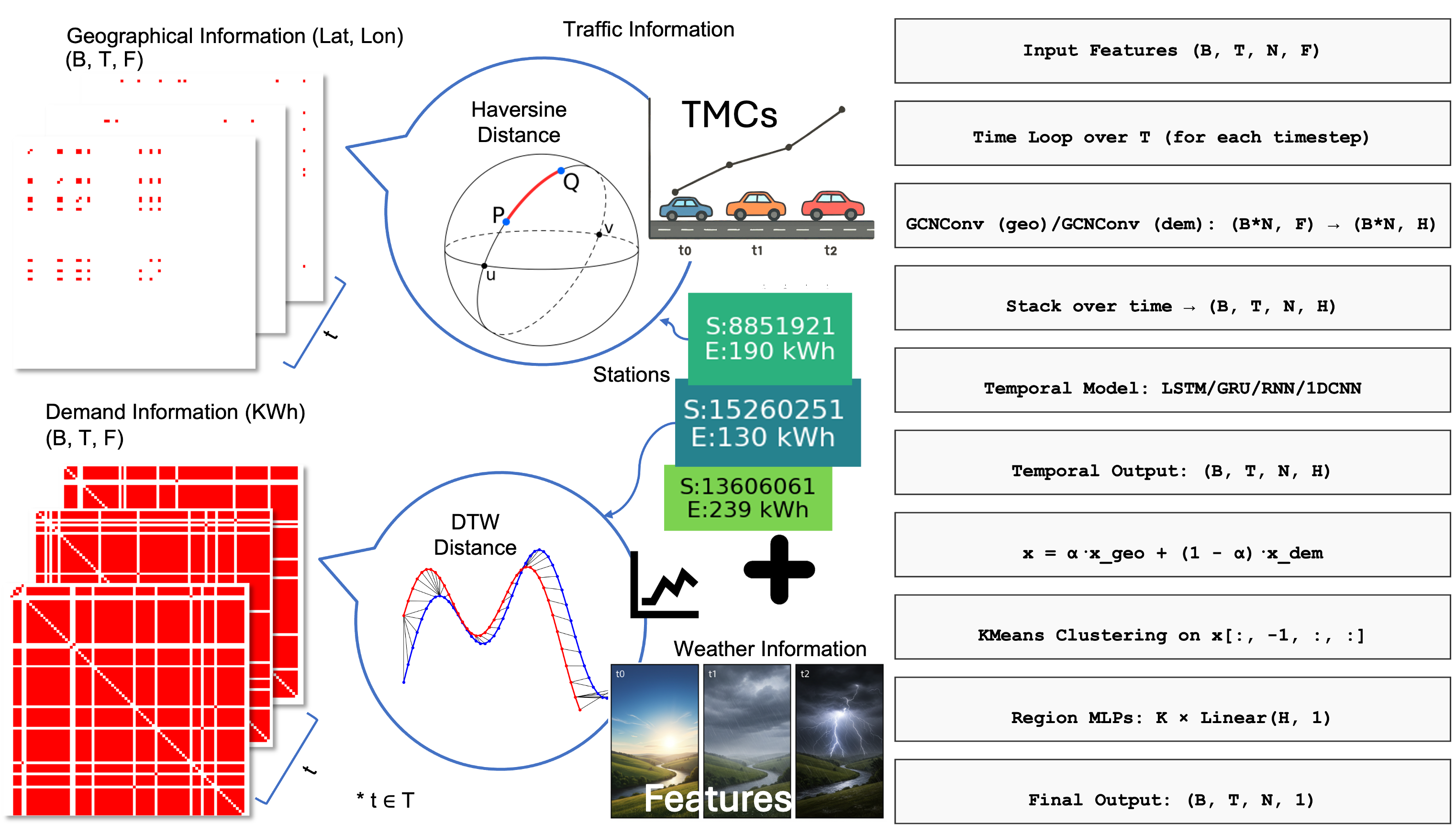}
\caption{The TW-GCN architecture processes spatiotemporal data (B, T, N, F) with added contextual features (weather, traffic, POIs). It employs GCN layers for spatial modeling at each timestep, transforming node features from (B$\times$N, F) to (B$\times$N, H), using Dynamic time warping (DTW) for temporal similarity and Haversine distance to transform arrays into adjacency matrices: Geographic (x\_geo) and demographic (x\_dem) features are later fused via a weighted sum ($\alpha$). The stacked outputs (B, T, N, H) feed into a temporal model. K-Means clustering on the final timestep's features identifies distinct regions, and K-MLPs generate region-specific predictions (B, T, N, 1), resulting in robust, region-aware forecasting.}
    \label{fig:enter-label}
\end{figure}

To effectively capture both spatial and temporal dependencies within the data, we construct relationships between nodes using two distinct adjacency matrices: one representing geographic proximity and another capturing the DTW similarity in demand. The geospatial adjacency matrix is defined as follows: given a set of \( N \) nodes, each associated with spatial coordinates \((\phi_i, \lambda_i)\) (latitude and longitude), we compute pairwise distances using the Haversine formula (Algorithm~\ref{alg:Haversine}). The resulting geospatial adjacency matrix \( A^{\text{geo}} \in \mathbb{R}^{N \times N} \) is constructed by applying a threshold \( \tau \) to ensure geographically proximate nodes are strongly connected (see Figure~\ref{fig:enter-label} for an overview of the architecture). Similarly, the temporal similarity matrix quantifies the similarity of demand patterns by computing the DTW distance, as defined in Algorithm~\ref{alg:DTW}, between the historical demand time series of each pair of nodes (Equation~\ref{eq:dtw}).

\begin{algorithm}
\caption{Dynamic Time Warping (DTW)}\label{alg:DTW}
\begin{algorithmic}[1]
\Require Two sequences $s = [s_1, s_2, \ldots, s_n]$, $t = [t_1, t_2, \ldots, t_m]$
\Ensure \texttt{DTW\_dist} – the DTW distance between $s$ and $t$
\State Initialize array \texttt{DTW}[0..n, 0..m] with $\infty$
\State \texttt{DTW}[0, 0] $\gets 0$
\For{$i \gets 1$ \textbf{to} $n$}
    \For{$j \gets 1$ \textbf{to} $m$}
        \State $cost \gets d(s_i, t_j)$
        \State \texttt{DTW}[$i, j$] $\gets cost + \min(\texttt{DTW}[i-1, j], \texttt{DTW}[i, j-1], \texttt{DTW}[i-1, j-1])$
    \EndFor
\EndFor
\State \Return \texttt{DTW}[n, m] \Comment{Final DTW distance}
\end{algorithmic}
\end{algorithm}

\begin{algorithm}
\caption{Haversine Distance}\label{alg:Haversine}
\begin{algorithmic}[1]
\Require \texttt{lat1, lon1, lat2, lon2} – coordinates in degrees
\Ensure \texttt{d} – Haversine distance in kilometers
\State Convert \texttt{lat1, lon1, lat2, lon2} from degrees to radians
\State $\Delta\phi \gets \texttt{lat2} - \texttt{lat1}$
\State $\Delta\lambda \gets \texttt{lon2} - \texttt{lon1}$
\State $a \gets \sin^2\left(\frac{\Delta\phi}{2}\right) + \cos(\texttt{lat1}) \cdot \cos(\texttt{lat2}) \cdot \sin^2\left(\frac{\Delta\lambda}{2}\right)$
\State $c \gets 2 \cdot \text{atan2}(\sqrt{a}, \sqrt{1 - a})$
\State $R \gets 6371$ \Comment{Earth radius in kilometers}
\State \texttt{d} $\gets R \cdot c$
\State \Return \texttt{d}
\end{algorithmic}
\end{algorithm}

\begin{equation}
d^{\text{DTW}}_{ij} = \text{DTW}(X_i, X_j),
\label{eq:dtw}
\end{equation}

where \( X_i \) and \( X_j \) are the demand time series for nodes \( i \) and \( j \), respectively. The scaling hyperparameter \( \gamma \) can be used to adjust sensitivity to temporal differences.

Before inputting the adjacency matrices into the GCN, we apply normalization (Equation~\ref{eq:norm}):

\begin{equation}
\tilde{A} = D^{-\frac{1}{2}} A D^{-\frac{1}{2}},
\label{eq:norm}
\end{equation}

where \( D_{ii} = \sum_j A_{ij} \). This ensures stable training and balanced node contributions. GCN layers are then computed as (Equation~\ref{eq:gcnlayer}):

\begin{equation}
H^{(l+1)} = \sigma\left( \tilde{A} H^{(l)} W^{(l)} \right),
\label{eq:gcnlayer}
\end{equation}

where \( H^{(l)} \) is the input at layer \( l \), \( W^{(l)} \) is a learnable weight matrix, and \( \sigma \) is a non-linear activation. The structure of \( \tilde{A} \) (Figure~\ref{fig:enter-label}) (e.g., single-graph or fused) depends on the model design.

\subsection{Clustering}

We employ the \textit{K}-means clustering algorithm on station-level data to uncover hidden spatial patterns in charging behavior. Prior research demonstrates that clustering effectively identifies distinct user groups with similar behavioral patterns~\citep{nespoli2023user, genov2024forecasting}, particularly in energy charging habits~\citep{al2019review}. These clusters often reflect underlying geographic context, enabling more accurate forecasting and tailored strategies.

After modeling spatiotemporal dependencies with the temporal component, we apply \textit{K}-means clustering to its output. This step groups stations with similar proximity profiles to different POI and traffic-weather types, effectively capturing shared spatial behavior. The resulting cluster assignments serve as an intermediate feature layer before the final MLP classifier (see Figure~\ref{fig:enter-label}), enriching station representations with spatial context and improving the accuracy of demand forecasting and infrastructure planning.

\section{Computational Experiments} \label{Computational Experiments }

We present key insights derived from our experimental results, highlighting the predictive capabilities of different models across varying temporal and spatial configurations. The following sections detail the performance of GCN architectures at the station level, followed by a baseline analysis incorporating temporal trends and spatial features such as surrounding points of interest. Experiments are conducted using Ubuntu 24 acting as the central orchestrator. We use PyTorch and PyTorch Geometric to deploy the GCNConv. The underlying hardware comprises an Intel(R) Xeon(R) Platinum 8462Y+, 256 GB RAM, and a Tesla L40 GPU with 48 GB memory running CUDA 12.2.

 We employ several deep learning architectures to model temporal dependencies and nonlinear dynamics in EV charging demand. The \texttt{RNN} serves as a foundational model capable of capturing sequential dependencies through recurrent connections. Building upon this, the LSTM network introduces gating mechanisms to mitigate vanishing gradient issues and effectively learn long-term temporal relationships. Similarly, the GRUs simplify the LSTM structure while maintaining comparable performance by combining the forget and input gates. One-Dimensional Convolutional Neural Networks (1DCNNs) leverage convolutional filters to extract local temporal patterns from time-series data, offering computational efficiency and robust feature extraction. Together, these architectures form the core of our modeling framework, enabling a comprehensive exploration of temporal dynamics in EV charging demand. Finetuning parameter grid is found and optimization parameter are found under Table \ref{tab:hyperparam}.

\begin{table}[H]
\centering
\scriptsize
\renewcommand{\arraystretch}{1.25}
\caption{Summary of \texttt{TW-GCN} model parameters, input dimensions, and parameter statistics under optimal configuration.}
\resizebox{\textwidth}{!}{%
\begin{tabularx}{1.05\textwidth}{l l l X}
\toprule
\textbf{Hyperparameter} & \textbf{Type} & \textbf{Default} & \textbf{Description} \\
\midrule
\texttt{node\_features} & int & 18 & Number of dynamic input features per node. \\
\texttt{hidden\_size} & int & 85 & Dimensionality of hidden representations for GCN and temporal layers. \\
\texttt{num\_regions} & int & 85 & Number of distinct geographic or functional regions. \\
\texttt{alpha} & float & 0.1 & Fusion weight between geographic and demand graph outputs: $x = \alpha x_{\text{geo}} + (1-\alpha)x_{\text{dem}}$. \\
\texttt{num\_clusters} & int & 10 & Number of K-Means clusters for regional grouping. \\
\texttt{model\_type} & str & 'LSTM' & Temporal model type. Options: \{'GRU', 'LSTM', 'BiLSTM', 'RNN', '1DCNN'\}. \\
\texttt{gcn\_geo}, \texttt{gcn\_dem} & GCNConv & --- & Graph convolution layers for geographic and demand graphs. Input: $(F)$, Output: \texttt{hidden\_size}. \\
\texttt{temporal\_model\_geo}, \texttt{temporal\_model\_dem} & nn.Module & LSTM & Recurrent models for temporal embedding in geographic and demand subgraphs. \\
\texttt{region\_mlp} & nn.ModuleList & Linear() & Region-specific output heads for local predictions. \\
\texttt{relu} & Activation & ReLU() & Non-linearity ensuring non-negative outputs. \\
\texttt{n\_init} & int & 10 & Number of random initializations for K-Means clustering. \\
\texttt{batch\_first} & bool & True & Treats input tensors as (batch, time, features). \\
\texttt{kernel\_size} & int & 3 & 1D convolution kernel size (1DCNN temporal model). \\
\texttt{padding} & int & 1 & Padding size to preserve sequence length in 1DCNN mode. \\
\midrule
\multicolumn{4}{l}{\textbf{Model Dimensions and Input Settings per Fold}} \\
\texttt{T (time steps)} & int & 7021 & Total number of temporal steps. \\
\texttt{N (locations)} & int & 85 & Number of nodes or spatial regions. \\
\texttt{F (features)} & int & 18 & Number of dynamic input features per node. \\
\texttt{GCN output features} & int & 18 & Output dimensionality of graph convolutional embeddings. \\
\texttt{hidden\_state shape} & Tensor & [batch, 85, 85] & Intermediate representation for temporal modeling. \\
\texttt{batch size} & int & 32 & Number of samples processed per training batch. \\
\texttt{epochs} & int & 100 & Total number of training epochs | Subject to early stoping. \\
\texttt{learning rate} & float & 0.001 & Step size used by the optimizer for gradient updates. \\
\midrule
\multicolumn{4}{l}{\textbf{Input and Batch Tensor Shapes}} \\
\texttt{adj\_geo batch} & Tensor & [32, 4, 85, 85] & Geographic adjacency matrix batch shape. \\
\texttt{adj\_dem batch} & Tensor & [32, 4, 85, 85] & Demand adjacency matrix batch shape. \\
\texttt{feature batch} & Tensor & [32, 4, 85, 18] & Node-level feature input tensor. \\
\texttt{target batch} & Tensor & [32, 4, 85, 1] & Ground truth target tensor. \\
\midrule
\multicolumn{4}{l}{\textbf{Layer and Parameter Summary}} \\
\texttt{Total parameters} & int & 52,730 & Includes GCN, recurrent, and MLP layers. \\
\texttt{Trainable params} & int & 52,730 & All parameters are trainable. \\
\texttt{Input size} & MB & 8.64 & Memory footprint of model inputs. \\
\texttt{Forward/backward size} & MB & 31.22 & Memory usage during forward and backward propagation. \\
\texttt{Parameter size} & MB & 0.21 & Total memory used by model parameters. \\
\texttt{Estimated total size} & MB & 40.07 & Approximate runtime memory usage per batch. \\
\bottomrule
\end{tabularx}%
}
\label{tab:hyperparam}
\end{table}

We conduct an ablation study by varying lag hours with values 1, 3, and 6. This parameter delimits how many hours in advance our prediction will forecast, several geographic clusters (\texttt{NUMCLUSTERSLIST} with 5 and 10), models (\texttt{MODELSLIST} with \texttt{`LSTM'}, \texttt{`GRU'}, \texttt{`1DCNN'}, \texttt{`RNN'}) representing what models are used, sequence lengths to create the necessary time windows to (\texttt{SEQLENGTHLIST} with 4, 8, and 12), and alpha values (\texttt{ALPHALIST} with 0.33, 0.5, and 0.66) weight probability during the concatenation process. Boolean parameters include traffic, weather, and adjacency features (\texttt{TRAFFICFEATURE}, \texttt{WEATHERFEATURE}, \texttt{ADJACENCY} set to \texttt{True}), and the number of epochs (\texttt{NUMEPOCHS} set to 125). The stacked architecture~\citep{wang2023short} was developed following the methodological framework proposed by~\citet{wang2023predicting}. The design of the network layers and the tuning of hyperparameters were informed by these prior works to maintain methodological consistency and ensure optimal model performance. Building on this foundation, the architecture was further adapted and applied to a proprietary ChargePoint dataset representing the study region.

\subsection{Prediction Analysis}\label{tab:Baseline}

\begin{figure}[!htpb]
    \centering
    \includegraphics[width=1\linewidth]{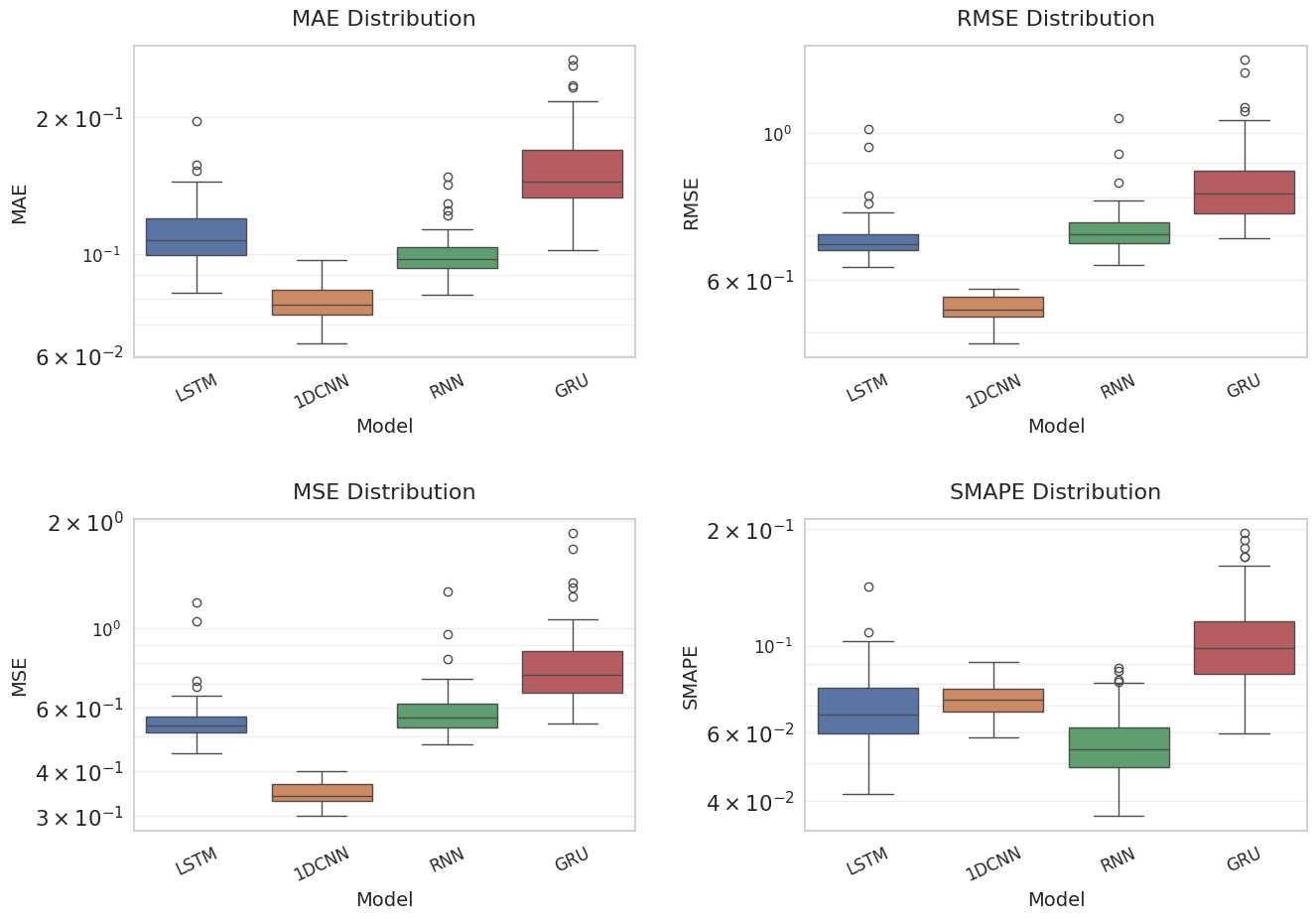}
    \caption{Distribution of error metrics (MAE, RMSE, MSE, SMAPE) across tuned deep learning models. Each boxplot is displayed on a logarithmic scale to highlight performance differences, with lower values indicating better accuracy.}
    \label{fig:dl_models}
\end{figure}

\begin{figure}[!htpb]
    \centering
    \includegraphics[width=1\linewidth]{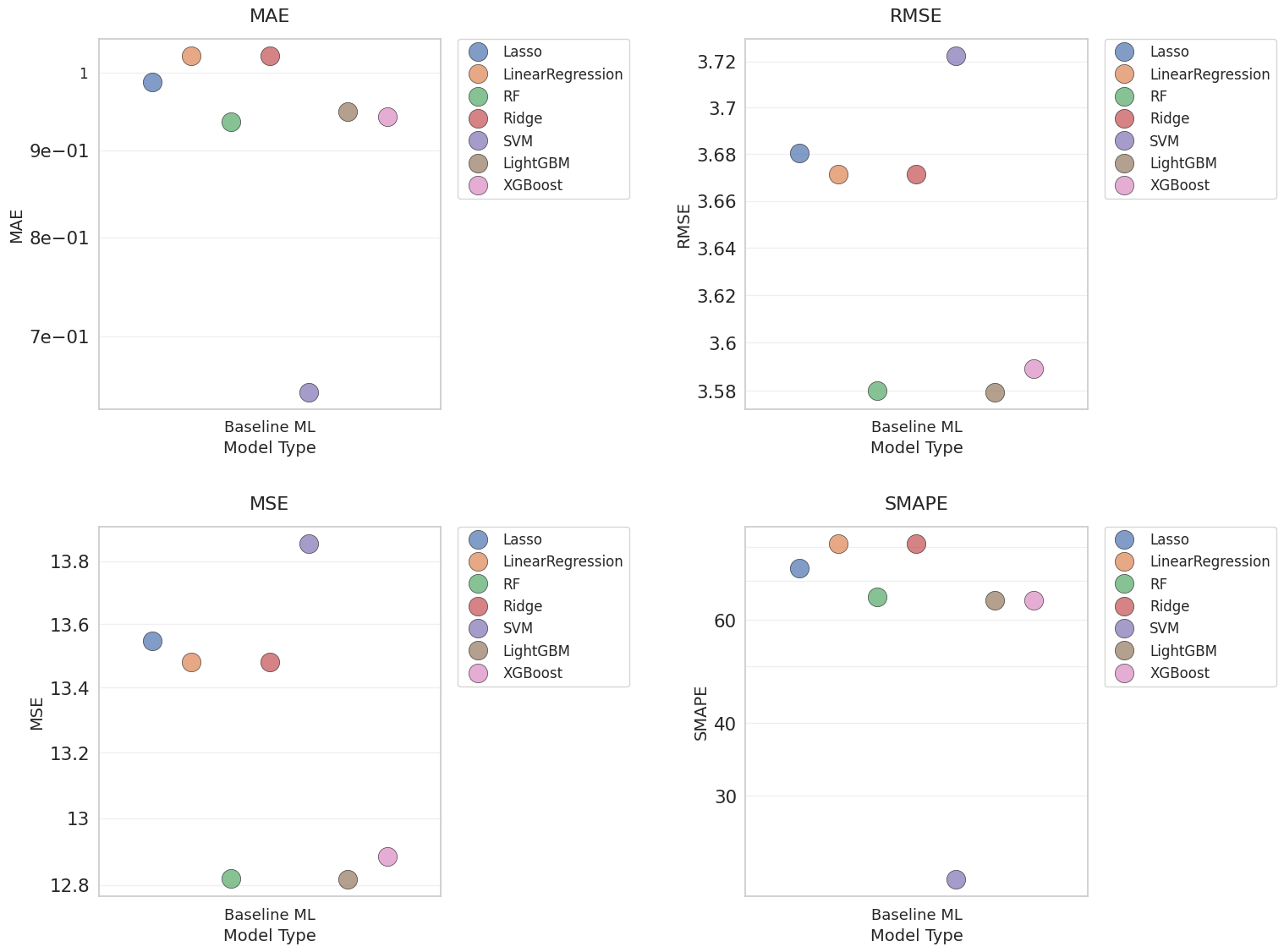}
    \caption{Comparison of TW-GCN models against baseline machine learning models across error metrics (MAE, RMSE, MSE, SMAPE). TW-GCN models are shown as boxplots, while baseline ML models are shown as overlaid stripplots for direct comparison.}
    \label{fig:twgcn_vs_ml}
\end{figure}

To evaluate the performance of each configuration\footnote{\url{https://github.com/XXXXXXXXX/XXXXXXXXX/blob/main/Readme.md} (MASKED FOR PEER REVIEW)}, we analyzed the average error metrics across combinations of lag hours, number of clusters, sequence lengths, and models. A visual summary of these relationships is provided in Figures~\ref{fig:dl_models}  and ~\ref{fig:twgcn_vs_ml}. %

Shorter lag hours, particularly LAG = 1, tend to yield lower MAE and RMSE for most models, as clearly shown in Figure~\ref{fig:dl_models}. For example, 1DCNN achieves its lowest MAE of 0.064 and RMSE of 0.575 at this lag. Increasing the number of clusters generally improves 1DCNN performance, whereas RNN and GRU models show more sensitivity to cluster size, with the optimal 1DCNN configuration corresponding to CLUSTERS = 10.

Moderate sequence lengths, typically between 4 and 8, often produce lower SMAPE values, indicating better relative forecasting accuracy; for instance, 1DCNN attains optimal performance with SEQ\_LENGTH = 8, suggesting that capturing recent temporal patterns without excessive historical context improves predictive reliability. Across all models, 1DCNN frequently outperforms LSTM, RNN, and GRU in terms of RMSE and MAE. Overall, the lowest MAE (0.064) and RMSE (0.573) are achieved by 1DCNN using the configuration LAG = 3, CLUSTERS = 5, and SEQ\_LENGTH = 8, suggesting that this setup best balances spatial clustering and temporal information. Full results extract can be found under (Tables~\ref{tab:tw-gcn} and~\ref{tab:base-models}) \footnote{Note: L = LAG\_HOURS, C = NUM\_CLUSTERS, $\alpha$ = ALPHA, SL = SEQ\_LENGTH}.

\begin{table}[htbp] 
\centering 
\tiny
\caption{Top 5 configurations per model type. (↓ indicates lower is better)}\label{tab:tw-gcn} 
\begin{tabular}{lcccccccc} 
\toprule 
\textbf{Model} & \textbf{L} & \textbf{C} & \textbf{$\alpha$ }  & \textbf{SL} & \textbf{MAE ↓} & \textbf{MSE ↓} & \textbf{RMSE ↓} & \textbf{SMAPE ↓} \\ 
\midrule 

1DCNN & 1 & 10 & 0.66 & 8 & \textbf{0.064} & 0.330 & 0.575 & \textbf{0.058} \\
 & 3 & 5 & 0.66 & 8 & \textbf{0.064} & \textbf{0.328} & \textbf{0.573} & 0.059 \\
 & 3 & 5 & 0.33 & 4 & 0.068 & 0.336 & 0.579 & 0.065 \\
 & 3 & 5 & 0.33 & 8 & 0.068 & 0.336 & 0.579 & 0.064 \\
 & 6 & 10 & 0.5 & 8 & 0.069 & 0.355 & 0.596 & 0.063 \\

\midrule 

RNN & 6 & 5 & 0.33 & 12 & \textbf{0.082} & 0.549 & 0.741 & 0.043 \\
 & 3 & 10 & 0.66 & 8 & 0.083 & 0.504 & 0.710 & \textbf{0.036} \\
 & 3 & 5 & 0.5 & 12 & 0.084 & 0.512 & 0.715 & 0.042 \\
 & 1 & 5 & 0.66 & 12 & 0.086 & 0.510 & 0.714 & 0.042 \\
 & 3 & 5 & 0.5 & 4 & 0.086 & \textbf{0.474} & \textbf{0.689} & 0.041 \\

\midrule 

LSTM & 3 & 5 & 0.33 & 12 & \textbf{0.082} & \textbf{0.464} & \textbf{0.681} & \textbf{0.042} \\
 & 6 & 5 & 0.5 & 8 & 0.088 & 0.508 & 0.713 & 0.050 \\
 & 6 & 10 & 0.5 & 8 & 0.092 & 0.509 & 0.714 & 0.057 \\
 & 1 & 5 & 0.66 & 4 & 0.093 & 0.470 & 0.685 & 0.052 \\
 & 1 & 5 & 0.66 & 12 & 0.094 & 0.520 & 0.721 & 0.050 \\

\midrule 

GRU & 3 & 5 & 0.33 & 12 & \textbf{0.102} & \textbf{0.544} & \textbf{0.738} & \textbf{0.060} \\
 & 1 & 10 & 0.33 & 8 & 0.116 & 0.553 & 0.743 & 0.071 \\
 & 1 & 10 & 0.66 & 12 & 0.117 & 0.583 & 0.763 & 0.081 \\
 & 1 & 5 & 0.66 & 8 & 0.121 & 0.605 & 0.778 & 0.074 \\
 & 3 & 10 & 0.33 & 8 & 0.124 & 0.628 & 0.793 & 0.080 \\

\bottomrule 
\end{tabular} 
\end{table}

\begin{table}[!htpb]
\centering
\tiny
\caption{Baseline models performance. (↓ indicates lower is better)}\label{tab:base-models}
\begin{tabular}{lcccc}
\toprule
\textbf{Model} & \textbf{MAE ↓} & \textbf{MSE ↓} & \textbf{RMSE ↓} & \textbf{SMAPE ↓} \\
\midrule
SVM & \textbf{0.649} & 13.857 & 3.722 & \textbf{21.564} \\
RandomForest & 0.936 & \textbf{12.819} & \textbf{3.580} & 65.710 \\
XGBoost & 0.941 & 12.885 & 3.589 & 64.790 \\
LightGBM & 0.948 & \textbf{12.816} & \textbf{3.579} & 64.911 \\
Lasso & 0.987 & 13.546 & 3.680 & 73.564 \\
LinearRegression & 1.022 & 13.479 & 3.671 & 81.178 \\
Ridge & 1.022 & 13.479 & 3.671 & 81.175 \\
\bottomrule
\end{tabular}
\end{table}

The analysis in Figures~\ref{fig:dl_models} and~\ref{fig:twgcn_vs_ml} highlights a clear separation between classical baselines and deep temporal–graph architectures. The boxplots illustrate that TW-GCN based models not only achieve consistently lower errors across all metrics but also display reduced variability compared to baseline methods. This robustness is especially relevant for practical deployment, where stable performance across different configurations is as critical as peak accuracy. In contrast, baseline machine learning models (Table~\ref{tab:base-models}) exhibit both higher errors and wider dispersion, reflecting their limited ability to capture nonlinear temporal and spatial dependencies inherent in the data.

Equally important is the comparative behavior among the deep learning architectures themselves, as summarized in Table~\ref{tab:tw-gcn}. While convolutional structures capture localized temporal dependencies, recurrent models demonstrate more nuanced sensitivity to hyperparameters such as lag depth and sequence length. Figure~\ref{fig:dl_models} further emphasize how these differences manifest across error scales. Together, these results suggest that the performance gap is not simply a matter of choosing the proposed TW-GCN over classical methods, but of selecting architectures that best exploit temporal and spatial dynamics.

\section{Analysis \& Insights} \label{Discussion}

Based on the results presented in Section  \ref{Computational Experiments }, we explore three optimized models and configurations to assess the impact of different temporal horizons (1, 3, 6) hours and clustering strategies on the predictive performance of EV charging station usage in Tennessee. For example, a 1-hour lag time (LAG\_HOURS = 1) combined with 10 clusters (NUM\_CLUSTERS = 10), a sequence length (SEQ\_LENGTH) of 4, and an alpha ($\alpha$) value of 0.66 can effectively capture immediate fluctuations. However, this configuration may also be overly sensitive to noise and transient anomalies, making it less reliable for modeling sustained behavioral trends. In contrast, 3- and 6-hour lags smooth out short-term volatility, better reflecting underlying patterns and enabling more stable predictions, critical for operational planning (e.g., shift scheduling, energy load balancing) and strategic decisions (e.g., infrastructure adjustments, demand forecasting).

A three hour lag increases the delay to 3 hours while reducing the number of clusters to 5, maintaining the same 1DCNN architecture, a sequence length of 4, and an $\alpha$ value of 0.66. Another configuration uses a 6-hour lag and reverts to 10 clusters, again employing the 1DCNN with a sequence length of 4 and $\alpha = 0.66$. Across all three models, the 1DCNN and sequence length remain consistent, suggesting a preference for this architecture in capturing temporal patterns, while variations in lag hours and cluster numbers explore different temporal and data partitioning strategies. The consistent alpha value indicates a stable weighting mechanism across configurations.

\subsection{Implications for Decision Makers}

The TW-GCN model places special emphasis on its ability to handle uncertainties in power grid behavior and accurately forecast electricity demand. In addition, a comparative analysis highlights TW-GCN's superior performance over traditional modeling approaches and variations in its internal structure. These alternatives often fail to capture key parameters such as points of interest, weather, traffic, and technological data limitations commonly observed in models like linear regression and decision trees.

Our study offers valuable insights for policymakers, urban planners, utility providers, and investors engaged in EV infrastructure planning. By utilizing TW-GCN, we enable accurate medium-term predictions of EV charging demand based on spatial, weather-related, and traffic variables. These predictive capabilities facilitate optimized infrastructure deployment, helping to prevent over or under investment in charging stations, particularly in slow adopting regions. 

Furthermore, the findings support more resilient power grid management by forecasting load distributions and peak usage times. For investors, this modeling improves decision making efficiency by identifying high-potential areas for deployment. The study also contributes to broader sustainability efforts by addressing the uneven spatial distribution of EV infrastructure, promoting equitable access, and supporting the reduction of transportation greenhouse gas emissions. Therefore, our approach can guide the creation of adaptive policies and incentive structures that are responsive to evolving patterns of demand and adoption. Subsequently, we examine how traffic conditions, weather, and POIs influence EV charging patterns, analyzing their relationships across the three main sources of information.


\subsubsection{Traffic}

By uniting the station dataset with the TMC traffic data, we were able to contextualize charging demand within the surrounding network. The nearest-neighbor matching approach allows each station observation to be associated with the traffic conditions of its closest road segments at the same point in time. This fusion revealed not only how traffic speed and travel time vary around charging locations, but also how reliability indicators such as delay per mile, confidence score, and reference speed influence accessibility. Analyzing these combined features provides key insights into the interaction between road congestion and charging station utilization to detect the importance of traffic-aware planning in understanding EV demand patterns. We compute the delay for each road segment relative to the reference speed. The analysis of traffic delays and EV charging behavior for January reveals that the overall correlation between delays and charging energy is weak, indicating that traffic congestion alone does not strongly influence charging demand.



However, a statistically significant difference in charging patterns is observed across delay segments. Low-delay stations record higher average energy consumption (6.45 kWh), while high-delay stations show substantially lower averages (4.76 kWh), with a mean difference of 1.69 kWh ($p<0.001$). This suggests that in congested areas, drivers tend to engage in shorter, opportunistic charging sessions rather than longer, high-capacity charges. From an executive perspective, infrastructure investments could prioritize larger-capacity charging hubs in low-delay zones, where demand is stronger and more consistent, while in high-delay zones, strategies should emphasize fast chargers designed for shorter sessions and higher turnover. While traffic delays are not a direct predictor of charging demand, they affect driver behavior and charging patterns. In our framework, this information forms part of the input to the stacked model used in the TW-GCN, enabling the network to better capture the interplay between traffic conditions and charging behavior in January and to guide resource allocation more efficiently.

A total of 2,522 charging events were analyzed to assess patterns of energy consumption and traffic delays at EV charging stations. The average energy consumption was 6.10 kWh, while the average traffic delay was 1.65 km/h. Observed energy consumption ranged from up to 64.79 kWh, and delays varied from -4.60 km/h to 24.36 km/h, indicating considerable heterogeneity in station usage and surrounding traffic conditions. The correlation between energy consumption and traffic delay was found to be very weak ($r = 0.067$), suggesting that while delays might slightly influence charging behavior, other factors such as station location, accessibility, and user habits are likely more important drivers of energy demand.  

Spatial analysis revealed stations can form up to 29 geographic clusters, with only two stations being isolated. This clustering highlights the concentration of charging infrastructure in high-demand areas, likely reflecting population density and urban planning considerations. When categorizing stations based on performance, 468 events of 2,520 were identified as high-energy, low-delay (optimal), whereas 317 events were high-energy, low-energy (inefficient). Optimal charging behavior is typically situated in locations that facilitate quick and frequent charging, while inefficient charging operations may face accessibility issues or lower utilization.  

Regional differences are also observed: the northern region, with 1,237 charging events, exhibited higher average energy consumption (7.56 kWh) and slightly higher delays (2.07 km/h), whereas the southern region, with 1285 stations, showed lower energy consumption (4.69 kWh) and shorter delays (1.24 km/h). This suggests that northern stations experience higher local demand and potentially longer dwell times. Analysis of station proximity further revealed that among station pairs separated by less than 5.5 km, the average energy difference was only 3.80 kWh, indicating consistent usage patterns among nearby stations and supporting the potential for coordinated load management in clustered areas.


These findings indicate that energy consumption at EV charging stations is only weakly related to traffic delays but strongly influenced by spatial clustering and regional demand patterns. Stations performing optimally tend to be located in high-demand areas with minimal congestion, and nearby stations exhibit similar energy usage, highlighting opportunities for targeted infrastructure planning and efficiency improvements.

\subsubsection{Weather}

\begin{figure}[!htbp]
    \centering
    \includegraphics[width=1\textwidth]{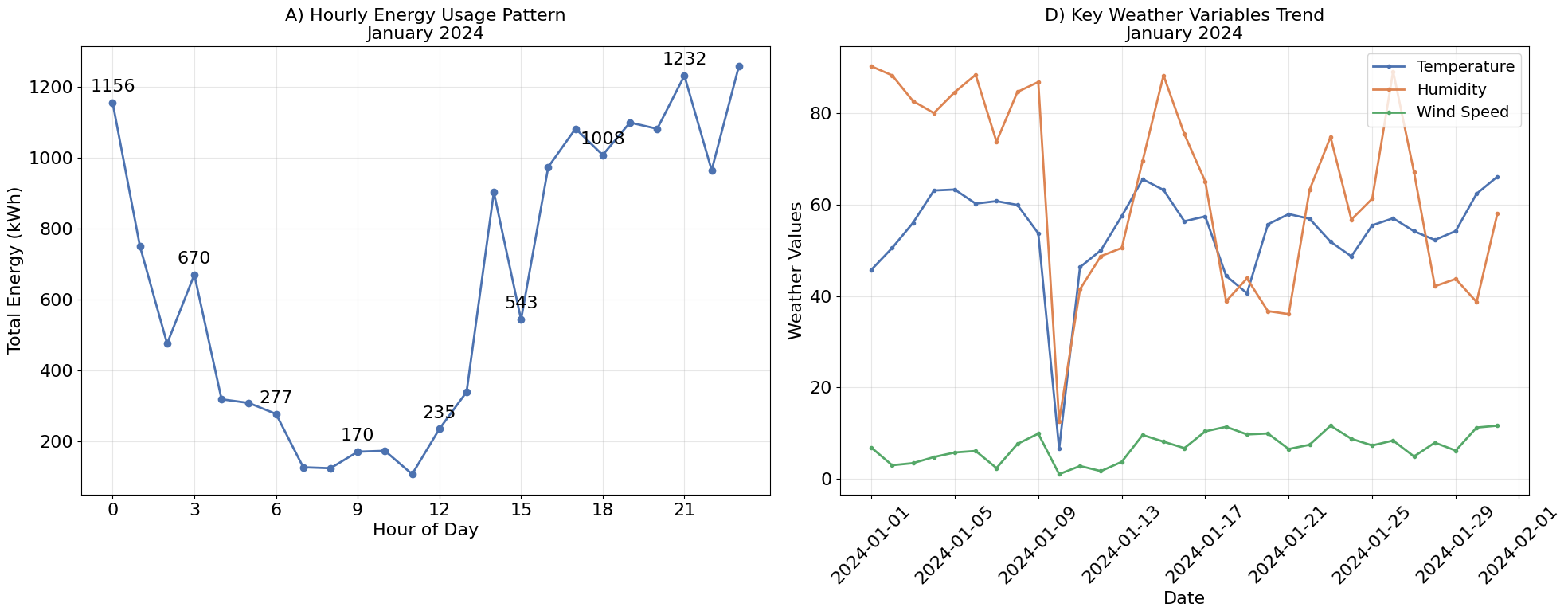} 
    \caption{Geographic distribution and energy consumption of EV charging stations for January 2024. The map visualizes 61 unique stations with 2,522 recorded charging events, highlighting spatial patterns of energy usage. Average energy per record is 6.10 kWh, and total energy consumption across all stations is 15,381 kWh. Weather conditions during this period are included for context, though correlations with energy usage are low, suggesting that EV charging demand is primarily driven by user behavior rather than weather factors.}
    \label{fig:energy_station_map}
\end{figure}

The analysis of EV charging activity for January 2024, shown in Figure~\ref{fig:energy_station_map}, covers 2,522 charging records and 1,939,365 weather observations from 45 weather stations. The average energy per charging event is 6.10 kWh, totaling 15,381 kWh for the month.

January, being a winter month, features variable conditions with temperatures ranging from -1°F to 78°F, humidity reaching up to 100\%, and wind speeds up to 32 mph. Correlation analysis indicates that energy demand is only weakly associated with weather variables, suggesting that EV usage is largely driven by human behavior rather than environmental conditions. However, daily weather patterns can still subtly influence charging: humid or rainy days may reduce trip frequency or encourage shorter charging sessions, while dry and mild days may promote longer trips and slightly higher energy consumption. Extremely cold days could slightly increase energy usage due to cabin heating or battery preconditioning. Overall, while weather introduces minor fluctuations, the primary determinants of energy consumption are user schedules, commuting patterns, and station availability throughout the studied period.

\subsubsection{Points of Interest}

Energy consumption was analyzed across various POIs, including hospitals, supermarkets, schools, restaurants, and parks. Overall, hospitals, schools, and parks exhibit similar statistics, with a mean energy consumption of approximately 6.10 kWh, a median of 1.84 kWh, and a standard deviation of 10.33 kWh across 2,522 observations, totaling around 15,381 kWh. Supermarkets show a more heterogeneous pattern: most charging operations (2,519) have a mean energy of 6.09 kWh, while a small subset of three locations exhibits higher consumption (9.68 kWh), suggesting potential high-consumption outliers. Restaurants also display variability, with the majority (2,498 locations) averaging 6.14 kWh, but a smaller group of 24 locations showing significantly lower usage (1.92 kWh). Statistical testing indicates that the presence of supermarkets is associated with higher mean energy consumption compared to locations without them, although this difference is not statistically significant ($t = 0.74$, $p = 0.54$). In contrast, locations with restaurants consume significantly less energy than those without ($t = -10.57$, $p < 0.001$). An interaction analysis between hospitals and supermarkets reveals that locations without both POIs (n = 2,519) have a mean energy of 6.09 kWh, while locations without hospitals but with supermarkets (n = 3) have a higher mean of 9.68 kWh, indicating a potential combined effect, though limited by small sample size. Overall, these results highlight that energy consumption patterns vary across POIs, with restaurants and supermarkets showing the most notable deviations.

\subsection{Prediction Horizon Analysis}

\begin{figure}[!htbp]
    \centering
    \includegraphics[width=0.95\linewidth]{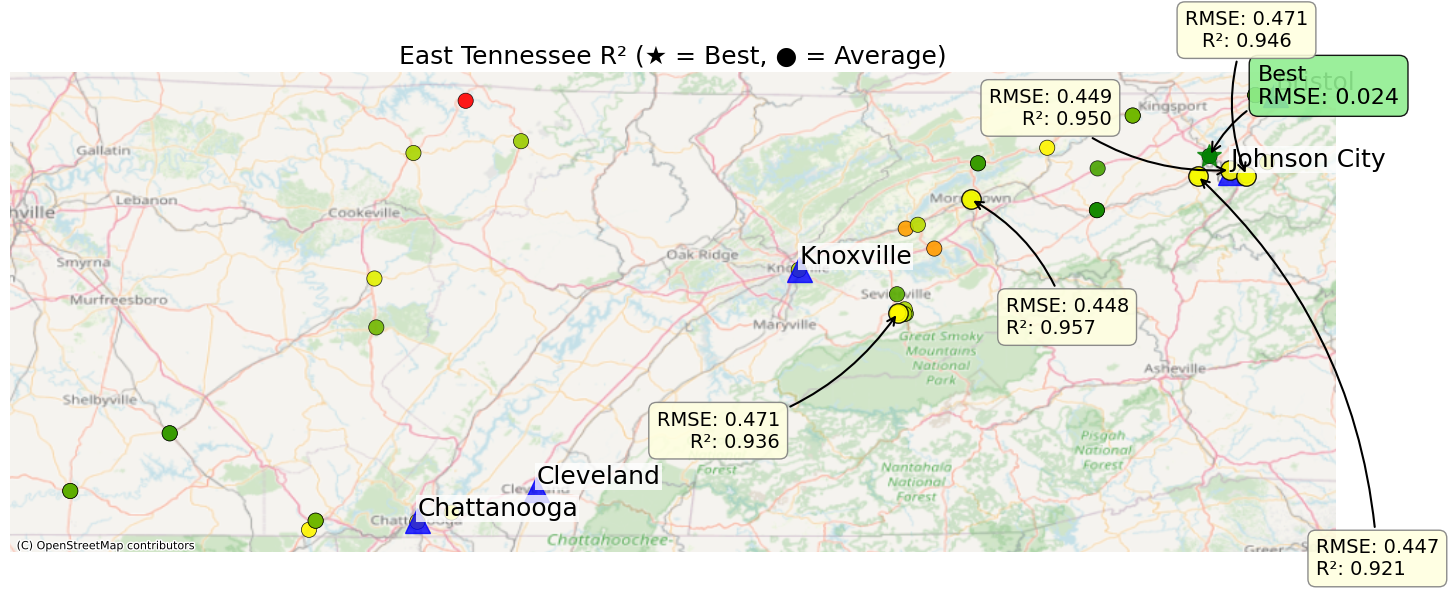}
    \caption{ Focused analysis of East Tennessee with best-performing station green star. Average stations are shown in yellow. Principal cities are labeled with blue triangles.}
    \label{fig:cap2}
\end{figure}

\begin{figure}[!htbp]
    \centering
    \includegraphics[width=0.95\linewidth]{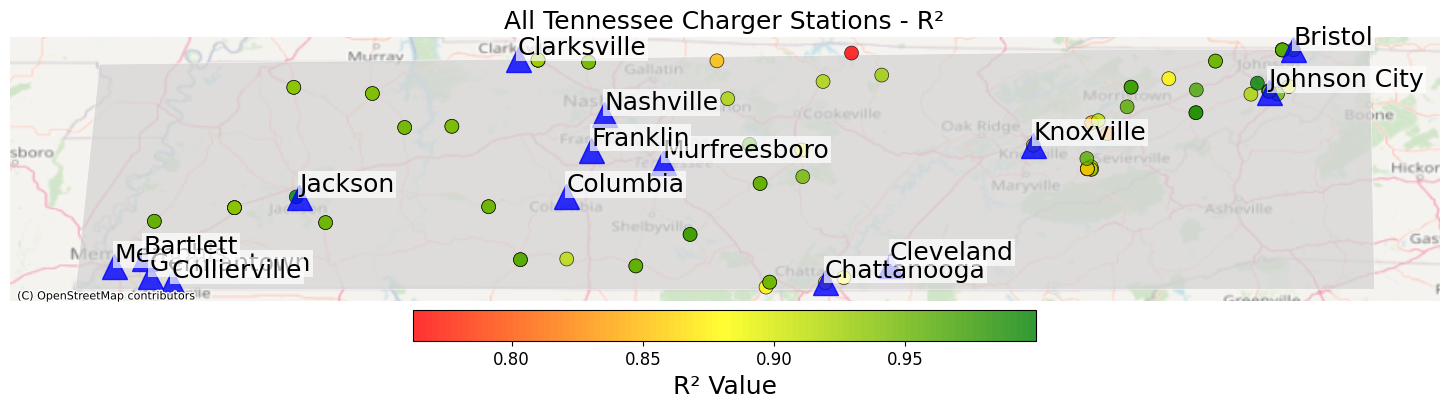}
    \caption{Performance analysis of individual charging stations using a 3-hour lag model and statewide view showing all stations colored by $R^2$ values.  }
    \label{fig:cap4}
\end{figure}

This section examines how the prediction horizon and regional characteristics affect model performance. By analyzing RMSE across Tennessee’s ZIP codes, we explore the trade-offs between responsiveness and stability in forecasts. Figure~\ref{fig:cap2} illustrates the geographic distribution of RMSE using a 3-hour lag model, which was identified as the best-performing configuration in our analysis. Key parameters for this model are: \(\texttt{LAG} = 3~\text{hours}\), \(\texttt{SEQ\_LENGTH} = 4\), and \(\texttt{CLUSTERS} = 10\). Our analysis of EV charging stations across Tennessee (see Figures~\ref{fig:cap2} and \ref{fig:cap4}) can guide new site deployments. By analyzing station-level and regional forecast performance metrics, including RMSE, $R^2$, and sample size, we identify high-confidence deployment regions and recommend a phased rollout strategy to maximize return on investment and minimize operational risk. We provide a strategic analysis of EV charging stations across Tennessee to guide a partnership with ChargePoint for new site deployments. The analysis focuses on identifying both high-performing and underperforming stations and regions, highlighting opportunities and risks for executive decision-making.

\begin{figure}[!htbp]
    \centering
    \includegraphics[width=1\linewidth]{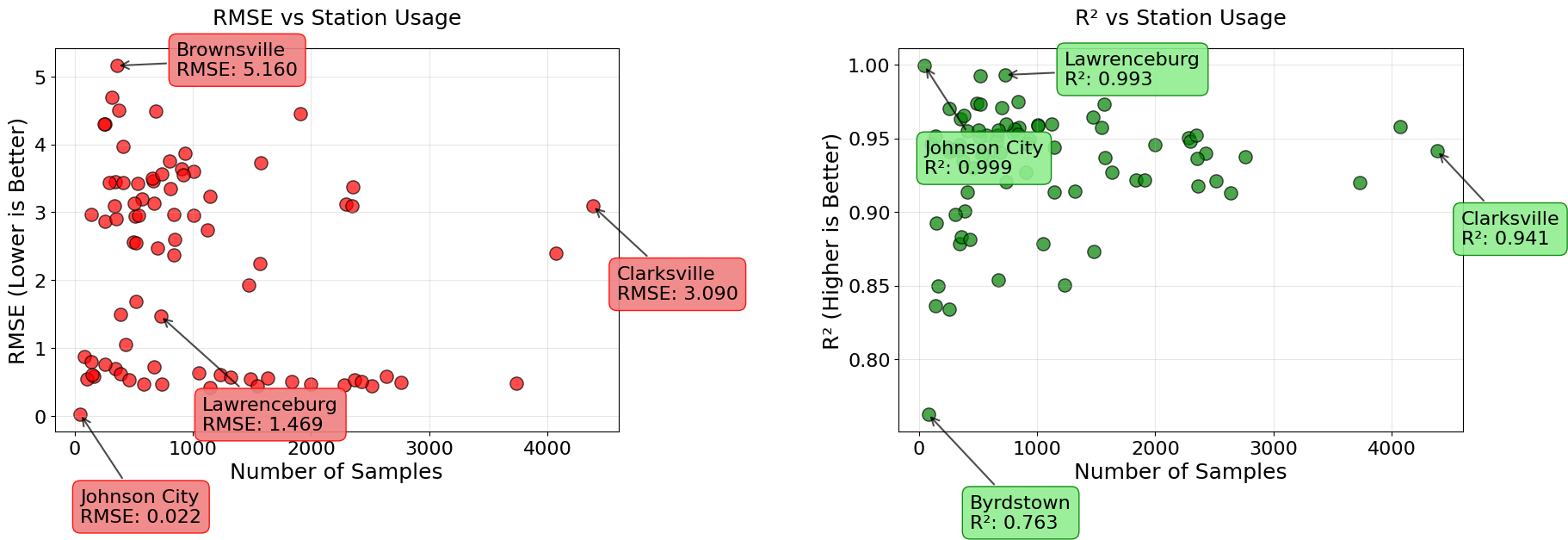}
    \caption{Relationship between station usage (number of samples) and forecasting performance metrics (RMSE, $R^2$, and MAE) across Tennessee EV charging sites. Each subplot illustrates how predictive accuracy varies with data availability. Lower RMSE and higher $R^2$ values are observed at high-usage stations, indicating improved model reliability as sample size increases.}

    \label{fig:cap3}
\end{figure}

\subsubsection{Regional Performance Insights}

East Tennessee demonstrates the most reliable forecast performance, with the lowest RMSE among all grand divisions. This indicates minimal absolute error in predicted station usage, making it the most suitable region for immediate deployment. West Tennessee, while showing the highest $R^2$, also exhibits the highest RMSE, reflecting substantial uncertainty in forecasted usage. Deployments in this region carry higher operational and financial risk, and additional data collection is recommended before committing to large-scale investment. Middle Tennessee displays moderate RMSE and strong $R^2$, suggesting that it represents a promising but cautious expansion opportunity following successful pilot deployments in East Tennessee.

\subsubsection{Station-Level Performance Insights}

Several stations stand out for exceptional performance. The BrightRidge site in Johnson City achieves near-perfect prediction metrics, with an RMSE of $0.022$ and an $R^2$ of $0.999$, indicating minimal operational uncertainty and high confidence in forecasted demand. Similarly, the Jackson site exhibits extremely accurate forecasts, ranking among the top performers with consistently low RMSE and high $R^2$ values, making it ideal for early deployment and partnership initiatives.
In contrast, a few stations display weaker predictive reliability. The Volunteer Parkway site in Bristol and a secondary BrightRidge location in Johnson City both present elevated RMSE and low $R^2$ values, suggesting substantial deviation between predicted and observed usage. These locations should be considered high-risk investments at present and warrant further observation or controlled pilot testing prior to full-scale deployment.
From a system-wide perspective, correlation analysis reinforces these insights. The relationship between RMSE and sample count is mildly negative ($r = -0.175$), consistent with the expectation that higher data availability improves model accuracy. Conversely, the correlation between $R^2$ and sample count is slightly positive ($r = 0.078$), indicating that stations with more historical data tend to exhibit more reliable forecasts.
Low-usage stations, which provide fewer observations, exhibit on average a $+28.9\%$ higher RMSE compared to high-usage stations (see Figure~\ref{fig:cap3}). This highlights the importance of sustained data collection at low-traffic sites to stabilize predictions over time. All stations passed quality screening, confirming that model variance stems primarily from behavioral rather than data-integrity factors.
Strategically, these results imply that high-usage, high-accuracy stations should serve as reference benchmarks for operational reliability, pricing calibration, and partnership modeling. Meanwhile, low-usage or high-variance sites should remain under watch or be integrated into limited pilot programs to validate demand potential before scaling investments. These findings, illustrated in Figure~\ref{fig:cap3}, provide a data-driven foundation for optimizing deployment sequencing, resource allocation, and collaborative expansion planning within Tennessee’s charging network.

\subsubsection{Strategic Deployment Recommendations}




\paragraph{Phase 1: Pilot Deployment} 
This phase should focus on the high-performing stations located in East Tennessee, where forecast reliability is the strongest. These sites present minimal operational risk and are ideal for testing early partnership models.

\paragraph{Phase 2: Secondary Expansion} 
The second phase should target stations in Middle Tennessee. Here, modular infrastructure can be deployed and scaled up gradually based on observed usage patterns and operational feedback.

\paragraph{Phase 3: Risk Assessment and Investigation} 
The final phase should include stations in West Tennessee and other low-performing sites. These areas require deeper analysis and targeted pilot projects to evaluate feasibility and mitigate risks before committing to full-scale investments.

\subsubsection{Operational and Business-Level Actions}

To optimize financial performance and operational efficiency, several strategic actions are recommended. Initial investments should prioritize high-performing stations characterized by low RMSE values and strong $R^2$ scores, ensuring reliable returns with minimal uncertainty. For moderate-performing sites, a modular deployment approach is advised to balance financial exposure and operational flexibility. Continuous monitoring of pilot sites will be essential to validate model forecasts and guide subsequent expansion decisions. Additionally, forming co-marketing or revenue-sharing partnerships at top-performing locations can enhance adoption and profitability. Investment in underperforming stations should be deferred until further historical data or improved predictive reliability becomes available.

\subsection{Business Level Decision}

ChargePoint providers can leverage predictive insights to make strategic, data-driven infrastructure decisions. Analysis of RMSE, population, and station density highlights underserved areas with latent demand, allowing providers to prioritize new station deployments where they are most likely to increase utilization. Business decisions should focus on regions where predictive models indicate gaps or high variability, ensuring investments are both efficient and aligned with actual usage patterns.
Operational efficiency can also be guided by predictive data. Forecasts from the 3-hour lag model enable providers to optimize energy allocation, maintenance scheduling, and staffing by anticipating peak usage periods. In regions with high RMSE or irregular demand patterns, data-driven interventions such as dynamic pricing, reservation systems, or temporary/mobile stations can smooth utilization and improve service reliability. These measures reduce wasted capacity and improve return on investment while ensuring stations are available where and when they are needed most.
Finally, data-driven customer engagement and growth strategies are essential for business-level decision-making. Providers can use predictive insights to offer real-time availability notifications, incentivize charging in low-utilization areas, and continuously refine deployment strategies based on actual usage data. Partnerships, expansion plans, and resource allocation should all be informed by the combination of RMSE, population density, and station performance metrics, enabling EV providers to expand efficiently, serve underserved markets, and maintain high customer satisfaction across the network.

\section{Conclusion \& Recommendations}\label{Conclusion}

This study introduces TW-GCN, a specialized spatio-temporal framework designed to enhance the prediction of electric vehicle charging demand across Tennessee. By fusing temporal modeling (1DCNN, LSTM, GRU, and RNN) with spatial dependencies captured via GCNs, the architecture effectively learns both temporal sequences and regional interconnectivity. Through extensive experimentation with real-world EV charging data and multiple model configurations, including variations in lag hours, clustering strategies, and prediction horizons. We demonstrate the model's capacity to balance short-term responsiveness with long-term stability. Key findings indicate that regional disparities in station availability and population density, especially in East Tennessee, can be meaningfully addressed by leveraging predictive performance at the ZIP or coordinates level. These insights equip decision makers with actionable intelligence for infrastructure deployment, policy alignment, and equitable EV accessibility.

 Our integrated use of weather, traffic, and spatial POIs further enhances the realism and applicability of the forecasting outputs, leading to a more robust modeling pipeline. Importantly, the short, mid, and long horizon analyses show that trade-offs between volatility and trend awareness must be carefully managed, with the mid horizon offering a particularly balanced outcome in terms of error performance and operational relevance.

 Building upon the results presented, future work could expand the dataset to include additional states or regions with different EV adoption rates can validate the model’s generalizability. Exploration of transformer-based architectures or attention mechanisms could enhance the ability to capture long-range dependencies across time and space. From a policy and infrastructure standpoint, integrating cost-benefit analysis into the modeling pipeline would support more precise investment decisions. Finally, addressing data gaps particularly and enhancing the granularity of station-level features (e.g., charging speed, stall availability) will further elevate the practical utility of the proposed approach.

 We note that the statements and findings presented in this study are solely those of the authors and do not necessarily reflect the views or positions of ChargePoint, Inc.